\documentclass[pdflatex,sn-mathphys-num]{sn-jnl}
\usepackage[T1]{fontenc}

\usepackage{graphicx}%
\usepackage{multirow}%
\usepackage{amsmath,amssymb,amsfonts}%
\usepackage{amsthm}%
\usepackage{mathrsfs}%
\usepackage[title]{appendix}%
\usepackage{xcolor}%
\usepackage{textcomp}%
\usepackage{manyfoot}%
\usepackage{booktabs}%
\usepackage{algorithm}%
\usepackage{algorithmicx}%
\usepackage{algpseudocode}%
\usepackage{listings}%
\usepackage{xurl}

\theoremstyle{thmstyleone}%
\theoremstyle{thmstyletwo}%

\theoremstyle{thmstylethree}%

\begin{document}

\title[Gradient-Based Evaluation of Sensitivity Fidelity]{Sensitivity Fidelity Reveals Physical Dependencies Beyond Forecast Accuracy in Deep-Learning Earth-System Models}


\author[1]{\fnm{Qi} \sur{Chen}}\email{qi.chen@kaust.edu.sa}
\author*[2]{\fnm{Peng} \sur{Zhan}}\email{zhanp@sustech.edu.cn}
\author[1]{\fnm{Hongxing} \sur{Cui}}\email{hongxing.cui@kaust.edu.sa}
\author[3]{\fnm{Wenqing} \sur{Cui}}\email{wenqing.cui@kaust.edu.sa}
\author*[1]{\fnm{Ibrahim} \sur{Hoteit}}\email{ibrahim.hoteit@kaust.edu.sa}

\affil*[1]{\orgdiv{Physical Science and Engineering Division}, \orgname{King Abdullah University of Science and Technology}, \orgaddress{\city{Thuwal}, \country{Saudi Arabia}}}

\affil*[2]{\orgdiv{Department of Ocean Science and Engineering}, \orgname{Southern University of Science and Technology}, \orgaddress{\city{Shenzhen}, \country{China}}}

\affil[3]{\orgdiv{Computer, Electrical and Mathematical Science and Engineering Division}, \orgname{King Abdullah University of Science and Technology}, \orgaddress{\city{Thuwal}, \country{Saudi Arabia}}}


\abstract{Deep-learning models increasingly rival numerical simulation in Earth-system forecasting, yet forecast error alone cannot determine whether their learned input--output dependencies are physically consistent. Here we introduce sensitivity fidelity as an additional criterion: alongside prediction error, model responses to input perturbations should be examined against a trusted reference or established physical knowledge. In an idealized Rayleigh--B\'enard convection benchmark, four surrogate architectures achieved low forecast errors yet encoded markedly different sensitivities. Quantitative comparison with a matched adjoint distinguished the models far more sharply than forecast accuracy and showed that predictive skill and sensitivity fidelity need not coincide, while finite-perturbation experiments further linked these differences to the propagation of initial-condition uncertainty. We then applied this strategy to El Ni\~{n}o--Southern Oscillation prediction, where no adjoint reference exists and learned sensitivities showed event-wise robustness, validity under finite perturbations, and consistency with established precursors. These results identify gradient-based evaluation as a discriminating complement to forecast-error metrics, providing a practical test of whether data-driven Earth-system models learn physically credible dependencies.}




\maketitle

\section{Introduction}\label{sec1}

Deep learning (DL) models have emerged as transformative tools in Earth science, providing data-driven representations of complex Earth system dynamics and computationally efficient surrogates for expensive numerical simulations~\cite{bodnar2025FoundationModelEarth,kochkov2024NeuralGeneralCirculation}. This unprecedented efficiency positions DL as a compelling alternative to traditional numerical modeling in operational settings~\cite{lam2023LearningSkillfulMediumrange, allen2025EndtoendDatadrivenWeather, price2025ProbabilisticWeatherForecasting}. However, the evaluation of these models relies almost exclusively on their ability to minimize prediction error, typically measured through data-misfit metrics such as mean-squared error (MSE)~\cite{liu2024EvaluationFiveGlobal}. While optimizing for these objectives yields high accuracy, it provides insufficient evidence that a model has learned the governing physical mechanisms~\cite{reichstein2019DeepLearningProcess}. Strong predictive skill alone does not resolve the black-box nature of DL~\cite{mcgovern2019MakingBlackBox}: a model may reproduce target fields by exploiting statistical shortcuts rather than encoding physically meaningful or mechanistically consistent mappings~\cite{geirhos2020ShortcutLearningDeep,lapuschkin2019UnmaskingCleverHans}. This concern is especially acute in scientific settings, where models are expected not only to predict accurately, but also to support mechanistic understanding, reliable inference, and responsible application~\cite{zhu2025UnveilingSensitiveDecisive,dramsch2025ExplainabilityCanFoster,mcgovern2024ValueConvergenceResearch}.

Establishing whether a model's forecasts rest on physically meaningful dependencies requires probing how they respond to the input state. We therefore evaluate DL models through their learned sensitivity patterns. In Earth-system modeling, sensitivity analysis is a foundational tool for uncovering physical mechanisms. It maps the complex spatiotemporal linkages driving forecast evolution~\cite{kleist2005InterpretationStructureEvolution, zhan2018SensitivityStudiesRed, ancell2007ComparingAdjointEnsembleSensitivity} and underpins critical applications such as targeted observing~\cite{torn2008EnsembleBasedSensitivityAnalysis}, parameter estimation~\cite{zhao2008ParameterEstimationThreedimensional}, and data assimilation~\cite{griffith2000AdjointMethodsData}. For nonlinear numerical models, however, sensitivity estimation has historically required either large numbers of forward perturbation experiments or dedicated adjoint models, which compute gradients efficiently through backward integration but are technically demanding to construct and maintain. Moreover, the sensitivities they provide are local and first-order, so their practical utility diminishes for finite-amplitude perturbations and long, strongly nonlinear forecast horizons~\cite{errico1997WhatAdjointModel,lea2000SensitivityAnalysisClimate}. Neural networks avoid the burden of constructing a dedicated adjoint. Because trained DL models define differentiable computational graphs, gradients of forecast objectives with respect to input states can be obtained directly through reverse-mode automatic differentiation~\cite{baydinAutomaticDifferentiationMachine,paszke2017AutomaticDifferentiationPyTorch,simonyan2014DeepConvolutionalNetworks}. This structure makes it possible to extract the sensitivity patterns encoded by the learned forecast operator at low marginal cost~\cite{shen2023DifferentiableModellingUnify}. However, these gradients characterize the learned forecast operator, not the underlying physical system. Recent evidence has shown that machine-learning weather models can exhibit unphysical tangent-linear and adjoint behavior~\cite{tian2026EvaluatingMachineLearning}. The dynamical consistency of backpropagated sensitivities therefore should not be assumed without explicit validation. Directly evaluating this consistency provides a more discriminating test of whether predictive skill reflects physically meaningful dependencies rather than statistical shortcuts.

This question has immediate practical relevance because backpropagated sensitivities are already used to interpret skillful DL forecasts~\cite{chen2024MachineLearningModel,wang2024RoleSeaSurface} and as substitutes for purpose-built adjoints in variational data assimilation~\cite{solvik20254DVarUsingHessiana,xiao2024SelfcontainedDatadrivenGlobal}. Both applications require confidence that learned gradients faithfully represent physically meaningful perturbation responses, yet predictive skill alone provides no such guarantee~\cite{lapuschkin2019UnmaskingCleverHans,toms2020PhysicallyInterpretableNeural}. Direct validation of sensitivities from end-to-end DL forecast models against physics-based references, however, remains limited. Previous studies have assessed the physical plausibility of AI-weather sensitivities~\cite{bano-medina2025AreAIWeather} and identified tangent-linear and adjoint deficiencies relevant to data assimilation~\cite{tian2026EvaluatingMachineLearning}, but have not systematically quantified the relationship between sensitivity fidelity and forecast performance. Moreover, because these comparisons involve distinct DL and physics-based systems rather than like-for-like dynamical counterparts, observed discrepancies may reflect differences in represented processes, state variables, numerical discretization, resolution, initialization, or objective formulation rather than deficiencies in the learned gradients themselves. Finite-difference tests have also validated surrogate-derived parameter gradients in subsurface inverse modeling~\cite{xiao2021DeepLearningBasedAdjointState}, but do not address the full-state sensitivities governing forecast evolution. Prior work on derivative-supervised learning has further highlighted the limitations of state-only fitting and the benefits of incorporating derivative information for approximation and downstream applications~\cite{czarnecki2017SobolevTrainingNeural,behroozi2026SensitivityConstrainedNeuralOperators}. Together, these findings motivate a central evaluation challenge: determining whether architectures trained using state-prediction losses alone exhibit differences in sensitivity fidelity that conventional forecast metrics fail to reveal. A controlled benchmark is therefore needed in which multiple architectures trained on the same dynamics to low forecast error are quantitatively evaluated against a matched physics-based adjoint, thereby isolating sensitivity fidelity as a property of the learned models. 

When a matched physics-based adjoint is unavailable, complementary diagnostics provide alternative means of evaluating sensitivities without requiring an explicit reference sensitivity. Existing approaches provide several forms of indirect evidence: frameworks for evaluating model explanations assess the robustness of model-derived interpretations~\cite{bommer2024FindingRightXAI}, while nonlinear perturbation studies test whether identified sensitive regions influence forecasts in ways consistent with known dynamics~\cite{guo2026NonlinearExplainableMethod}. Taken together, these approaches suggest three complementary forms of evidence for evaluating learned gradients: statistical robustness across multiple cases~\cite{bommer2024FindingRightXAI}, quantitative agreement with finite-perturbation responses~\cite{torn2008EnsembleBasedSensitivityAnalysis,guo2026NonlinearExplainableMethod}, and consistency with established physical mechanisms~\cite{reichstein2019DeepLearningProcess,toms2020PhysicallyInterpretableNeural}.

\begin{figure}[htbp]
\centering
\includegraphics[width=1.0\textwidth]{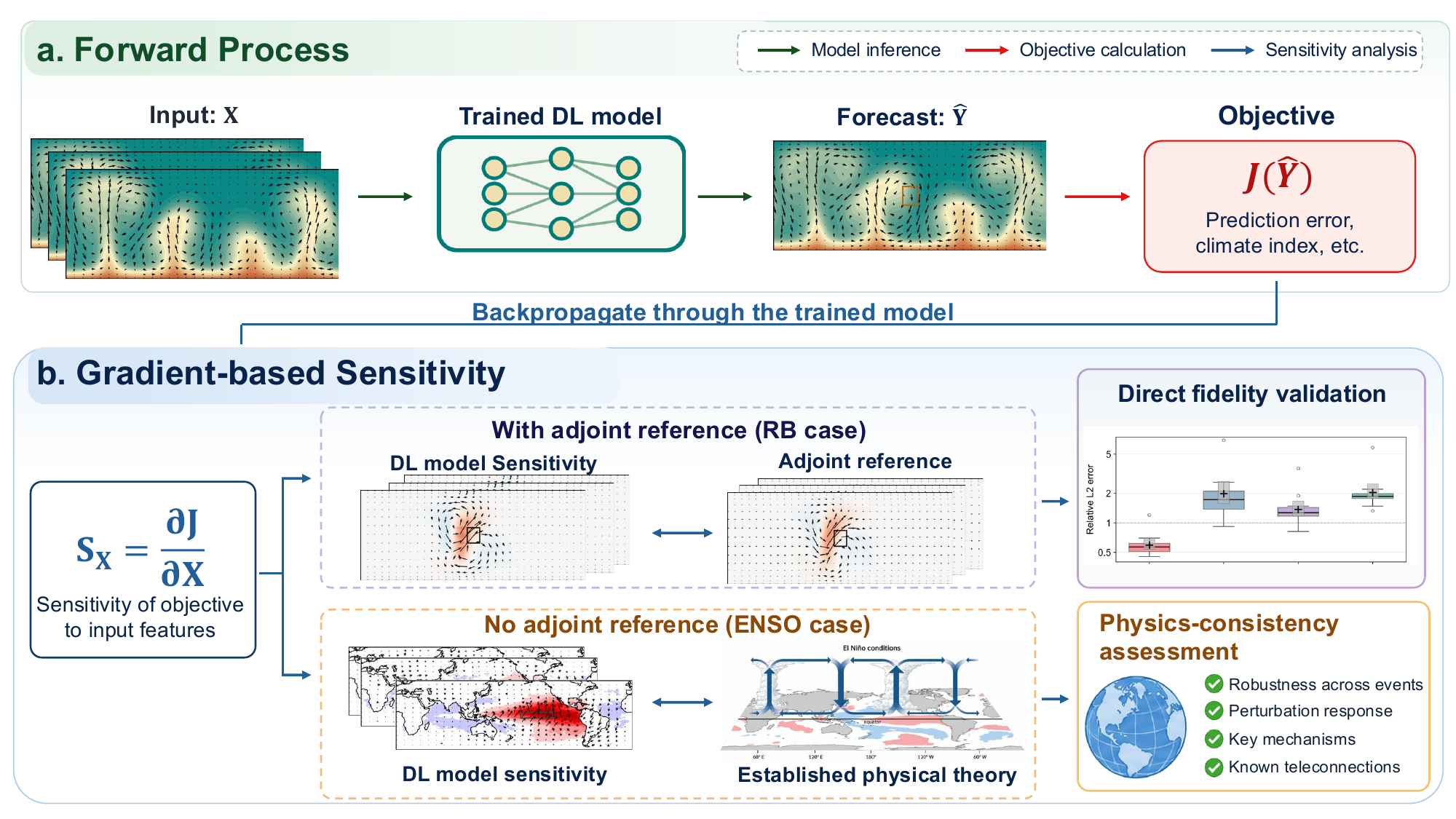}
\caption{\textbf{Sensitivity-fidelity workflow for differentiable Earth-system models}. (a) A trained DL forecast model maps an input state or input sequence $X$ to a forecast $\hat{Y}$, from which a scalar objective $J(\hat{Y})$ is selected. (b) Backpropagation through the trained model yields the input sensitivity $S_X=\partial J(\hat{Y})/\partial X$, which characterizes the learned local, first-order response of the objective to input perturbations. In the RB benchmark, neural sensitivities are directly compared with sensitivities from a matched physics-based adjoint for fidelity validation. In the ENSO application, where no adjoint reference is available, sensitivities are assessed through statistical robustness across events, quantitative agreement with the model's own finite-perturbation responses, and consistency with established physical mechanisms. These diagnostics provide complementary indirect evidence for evaluating learned sensitivities.}
\label{fig:sensitivity-framework}
\end{figure}

Here we introduce a sensitivity-fidelity framework for evaluating differentiable forecast models through both predictive accuracy and the physical faithfulness of their gradients (Fig.~\ref{fig:sensitivity-framework}). The framework generates forecasts with a trained differentiable model and backpropagates a scalar forecast objective to obtain sensitivities to the inputs. We test the framework in two settings spanning controlled dynamics and realistic climate prediction. First, we consider idealized two-dimensional Rayleigh--B\'enard (RB) convection, a dynamically rich yet fully controlled system in which training trajectories are generated by direct numerical simulation and reference sensitivities by a matched physics-based adjoint. We obtain sensitivities by backpropagating through four surrogate architectures trained on the same trajectories, encompassing Fourier-operator and attention-based designs. Quantitative comparison with the adjoint reference tests whether models with low forecast errors also recover the sensitivity structure of the underlying dynamics, and a perturbation-ensemble experiment extends this assessment to the propagation of initial-condition uncertainty. Second, we apply the same gradient diagnostics to El Ni\~{n}o--Southern Oscillation (ENSO) prediction, a realistic climate setting without an independently trusted physics-based adjoint. Instead, learned sensitivities are assessed through statistical robustness across El Ni\~{n}o events, quantitative agreement with the model's finite-perturbation responses and consistency with established ENSO dynamics. Together, these cases demonstrate how sensitivity fidelity can be benchmarked directly when a matched adjoint is available and examined through converging indirect evidence when it is not. Evaluating sensitivity fidelity alongside predictive accuracy therefore provides a discriminating complement to forecast-error metrics and a prerequisite for the reliable use of learned gradients. More broadly, this framework marks a step toward data-driven Earth-system models that can be trusted not only for what they predict, but also for how they respond to perturbations.

\section{Results}\label{sec2}

\subsection{Rayleigh--B\'enard convection benchmark for sensitivity fidelity}

A central question for differentiable surrogate models is whether accurate forecasts imply that the model has learned the physical dependencies governing the system. We investigate this question using two-dimensional RB convection, whose coupled temperature--velocity dynamics are sufficiently rich to challenge neural operators while still permitting reference sensitivities computed with an adjoint model. The adjoint gradient provides a physics-based reference against which learned sensitivities can be evaluated. On identical RB trajectories, we train four differentiable surrogates that autoregressively predict the system states: FNO~\cite{li2021FourierNeuralOperator}, AFNO~\cite{guibas2022AdaptiveFourierNeural, pathak2022FourCastNetGlobalDatadriven}, Swin Transformer~\cite{liu2022SwinTransformerV2a}, and IFactFormer-RB~\cite{li2023ScalableTransformerPDE,yang2025ImplicitFactorizedTransformer}. After training, each surrogate backpropagates a fixed target-region forecast metric to the initial states, producing sensitivity fields that are compared with the adjoint reference. This comparison tests whether forecast accuracy is accompanied by physically consistent gradients.

\begin{figure}[b]
\centering
\includegraphics[width=1.0\textwidth]{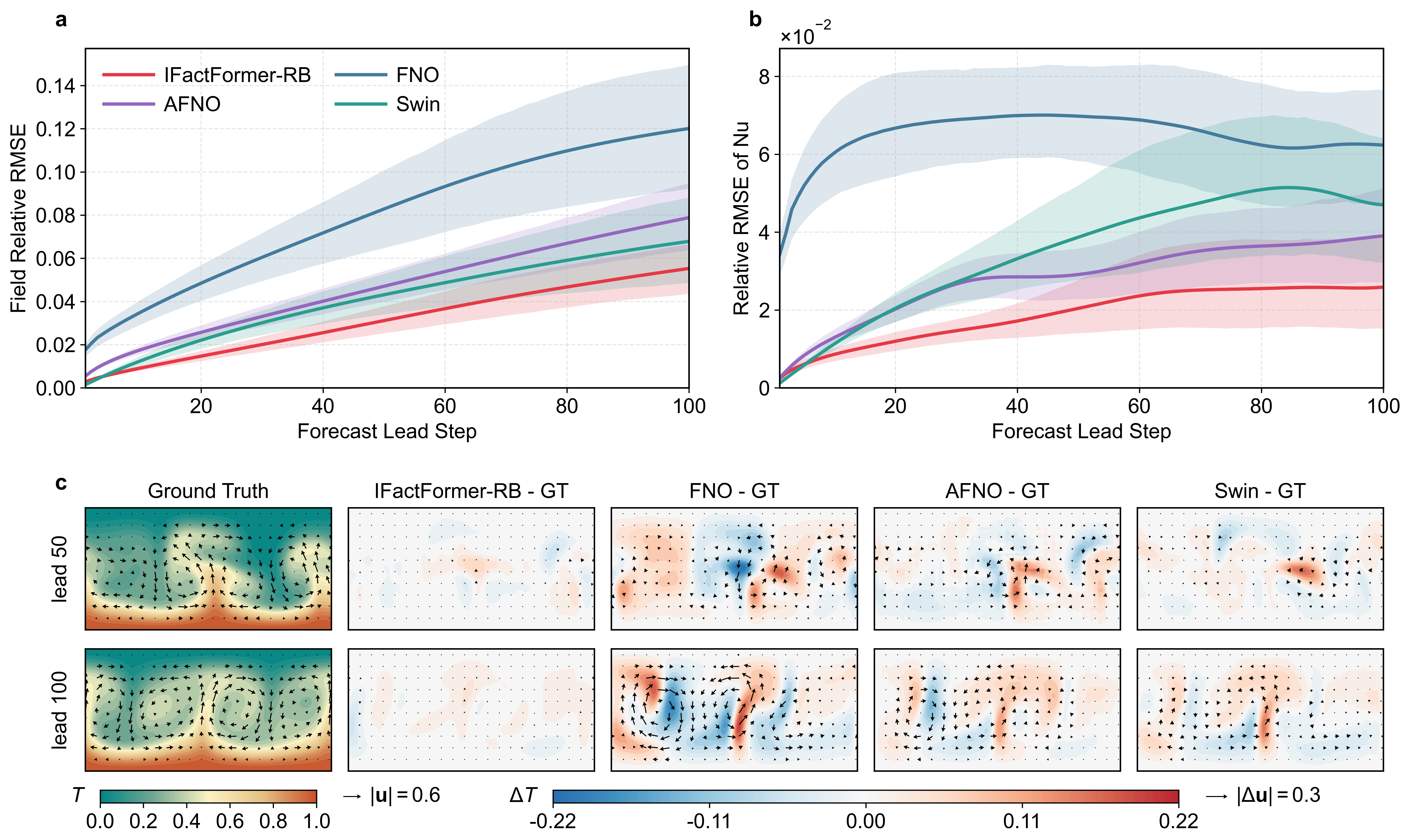}
\caption{\textbf{Forecast performance of different surrogate models for two-dimensional RB convection}. (a) Full-field relative root-mean-square error (RMSE) on the held-out test set as a function of forecast lead step for IFactFormer-RB, FNO, AFNO, and Swin Transformer. (b) Corresponding relative RMSE of the predicted Nusselt number. In (a) and (b), colors distinguish the four models, solid lines show the mean relative RMSE across held-out test trajectories at each forecast lead step, and shaded bands indicate the corresponding 95\% bootstrap confidence intervals. (c) Ground-truth states and forecast residuals for a representative held-out trajectory at lead steps 50 (top row) and 100 (bottom row). The first column shows the ground-truth temperature and velocity fields; the remaining columns show predictions minus ground truth for IFactFormer-RB, FNO, AFNO, and Swin Transformer, respectively. Color shading denotes temperature $T$ or temperature residuals $\Delta T=\hat{T}-T$, and arrows denote velocity $\mathbf{u}$ or velocity-vector residuals $\Delta\mathbf{u}=\hat{\mathbf{u}}-\mathbf{u}$, with separate color scales and reference arrows for the ground-truth and residual fields.}
\label{fig:rb-forecast}
\end{figure}

All four surrogates reproduce the evolution of buoyant plumes and the associated recirculating cells throughout the rollout (Fig. S1). Across held-out trajectories, the mean full-field relative RMSE increases gradually over the 100-step prediction horizon but remains below approximately 0.12 for every model (Fig.~\ref{fig:rb-forecast}a). IFactFormer-RB is consistently the most accurate, with errors remaining below 0.06 even at the final lead step, followed by Swin Transformer and AFNO, while FNO exhibits the greatest error accumulation. This performance ranking is also evident in the temperature and velocity residuals at lead steps 50 and 100 for a representative trajectory (Fig.~\ref{fig:rb-forecast}c). Thus, judged solely by full-field forecast error, all four surrogates produce stable long-horizon rollouts, with only moderate differences in accuracy. To assess physically relevant bulk behavior beyond full-field agreement, we next turn to the Nusselt number, an integral measure of convective heat transport (Fig.~\ref{fig:rb-forecast}b). This diagnostic yields a partly different model ranking: the Nusselt-number error of FNO increases rapidly before plateauing near 0.07, whereas that of IFactFormer-RB remains below 0.03 throughout the rollout. Notably, although Swin Transformer outperforms AFNO in full-field accuracy, its Nusselt-number error grows steadily and eventually exceeds that of AFNO. The same behavior occurs in the Nusselt-number evolution of a representative trajectory, in which the Swin Transformer prediction departs from the ground truth near the peak in convective transport (Fig. S1). This change in ranking indicates that full-field accuracy does not necessarily translate into a faithful representation of bulk heat transport. As a scalar summary of the flow, however, the Nusselt number can identify such aggregate discrepancies but cannot reveal the input--output dependencies learned by each model. We therefore turn to the gradient-based sensitivity fields encoded by the four surrogates, which reveal how their predicted future states depend on the initial condition and, as shown next, distinguish the models more sharply than either forecast diagnostic.

\begin{figure}[htbp]
\centering
\includegraphics[width=1.0\textwidth]{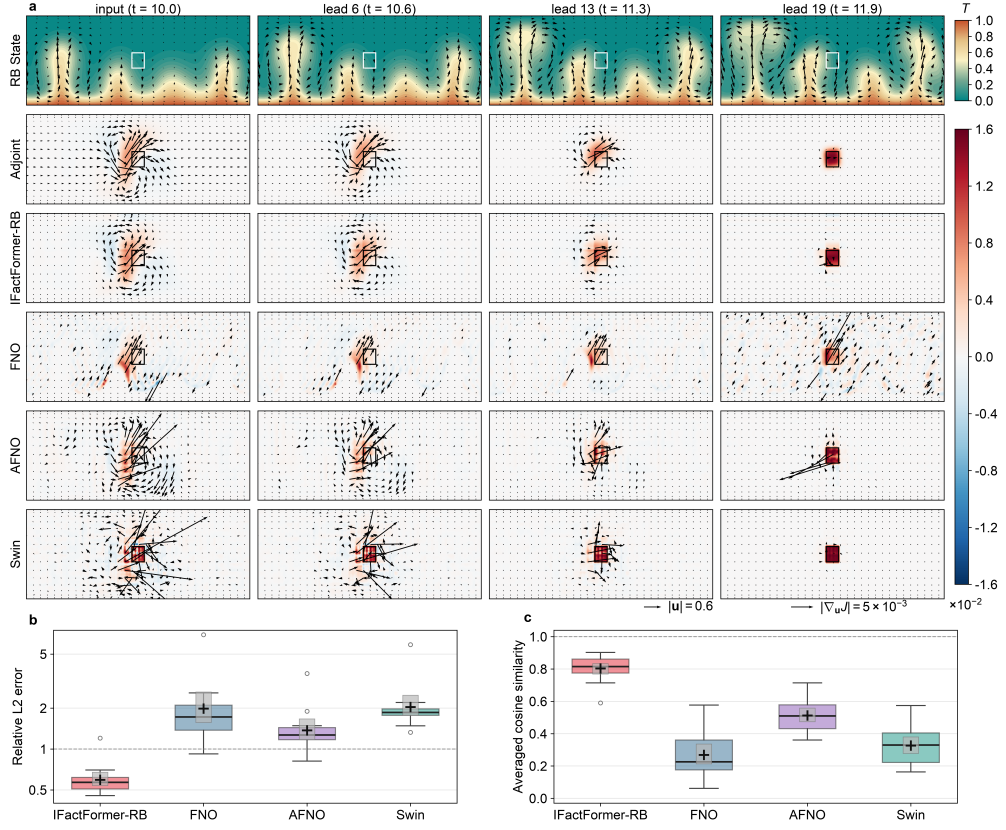}
\caption{\textbf{Comparison of surrogate sensitivity fields with the adjoint reference for the RB forecast metric}. The forecast metric is defined as the mean temperature within the target region at forecast lead step 20. (a) RB states and corresponding sensitivity fields for one held-out test trajectory. The top row shows the state at the input time ($t=10.0$) and at forecast lead steps 6, 13, and 19. Color shading represents temperature, and arrows represent velocity. The remaining rows show the corresponding sensitivity fields computed using the adjoint model, IFactFormer-RB, FNO, AFNO, and Swin Transformer. In these rows, color shading represents sensitivity to temperature, and arrows represent sensitivity to velocity. White boxes identify the target region in the state panels, and black boxes identify the same region in the sensitivity panels. (b) Relative $L_2$ error between each surrogate sensitivity field and the adjoint reference. (c) Averaged cosine similarity between each surrogate sensitivity field and the adjoint reference. In (b) and (c), boxplots show the distributions of the corresponding metrics across the held-out test trajectories. Crosses indicate sample means, and shaded bars indicate the 95\% bootstrap confidence intervals of the means.}
\label{fig:rb-sensitivity}
\end{figure}

The sensitivity comparison shows that this forecast reliability does not necessarily extend to the gradients encoded by the surrogates (Fig.~\ref{fig:rb-sensitivity}). Here, the forecast metric is the mean temperature within the target region at lead step 20, and each sensitivity field is the gradient of this metric with respect to the temperature and velocity fields at an earlier time. For the representative held-out trajectory, the adjoint sensitivity does not remain centered on the target region but propagates backward along the upstream dependency pathway of the RB flow (Fig.~\ref{fig:rb-sensitivity}a). At early times, the temperature sensitivity forms an elongated band displaced from the target and aligned with the flank of a convective structure that subsequently enters the target region. The corresponding velocity sensitivity forms a coherent vector pattern that identifies how perturbations can advect or deform this structure. As the objective time approaches, both components contract toward the target. This organized backward propagation provides a physics-based reference for evaluating the sensitivities learned by the surrogates. IFactFormer-RB reproduces this evolution most faithfully, including the displaced early-time temperature sensitivity, its subsequent contraction toward the target, and the organization of the associated velocity sensitivities. FNO recovers part of the positive temperature band but introduces spurious small-scale structure and diffuse velocity responses. AFNO approximately localizes the relevant region, although its gradients remain displaced and noisy. Swin Transformer provides the clearest contrast, despite achieving full-field forecast accuracy close to that of IFactFormer-RB over the same 20-step horizon. Its temperature sensitivity shifts toward the target but remains strongly fragmented. The continuous, predominantly single-signed band in the adjoint is replaced by mixed-sign patches, with exaggerated amplitudes near the target and little sensitivity along much of the reference pathway. Its velocity sensitivities are similarly dominated by scattered, large-amplitude vectors that lack the coherent organization of the adjoint reference.

The quantitative metrics confirm these differences (Fig.~\ref{fig:rb-sensitivity}b, c). The relative $L_2$ error measures the overall deviation of each surrogate sensitivity field from the adjoint in both amplitude and spatial structure (Fig.~\ref{fig:rb-sensitivity}b). Across the held-out test trajectories, the error for IFactFormer-RB remains below one for all but one trajectory and has a mean of approximately 0.6. By contrast, the distributions for the other three surrogates lie almost entirely above one, indicating deviations larger than the norm of the reference field itself. Their mean errors are approximately 1.4 for AFNO and close to 2 for FNO and Swin Transformer. Beyond this overall discrepancy, much of the physical information carried by a sensitivity field lies in its spatial organization, which identifies where the forecast metric depends on the initial condition and captures the sign and orientation of that dependence. We therefore also consider the averaged cosine similarity, which directly measures the agreement between each surrogate sensitivity field and the adjoint reference in this spatial pattern (Fig.~\ref{fig:rb-sensitivity}c). The same ordering emerges. IFactFormer-RB has a narrow distribution above 0.7 for all but one trajectory and a mean similarity of approximately 0.8. AFNO has a broader distribution with a mean near 0.5, while FNO and Swin Transformer both have mean similarities near 0.3, indicating only weak recovery of the adjoint pattern. In contrast to the moderate spread in forecast accuracy, the gap between models is far larger in sensitivity fidelity. Swin Transformer achieves the second-lowest full-field forecast error yet agrees with the adjoint no better than FNO, the least accurate forecaster. Additional experiments using different target regions and longer objective lead times show the same separation among the surrogate sensitivities (Figs. S2 and S3). We further test whether these differences persist for finite perturbations by propagating a prescribed perturbation ensemble through both the reference simulation and each surrogate (Text S3, Fig.~S4). Because the adjoint provides the reference gradient of the simulation, the discrepancy between its linear prediction and the corresponding nonlinear response defines a natural baseline for departures from the linear regime. This discrepancy remains small over most of the perturbation range and becomes appreciable only at the largest amplitudes. In parallel, for each surrogate, the response predicted from its sensitivity closely follows that obtained from direct rollout, confirming that the learned gradients reliably describe the surrogate's own local perturbation behavior. The remaining differences from the reference response therefore primarily reflect errors in the learned sensitivities rather than nonlinear effects. Consistent with the direct sensitivity comparison, these response errors remain well above the nonlinear-error baseline, and the model ranking is unchanged across perturbation amplitudes: IFactFormer-RB performs best, followed by AFNO, FNO and Swin Transformer. Thus, differences in sensitivity fidelity translate directly into differences in how initial-condition uncertainty is propagated. Accurate RB forecasts do not imply that a surrogate has learned the physical dependencies governing the forecast metric, making adjoint sensitivity comparison a more stringent test of dynamically consistent input--output dependencies.

\subsection{Evaluating ENSO precursor sensitivities without an adjoint reference}

Having established in the RB benchmark that forecast accuracy alone does not guarantee faithful sensitivities, we next consider how learned sensitivities can be evaluated in the absence of an adjoint reference. As a test case we use CTEFNet~\cite{chen2025LongrangeENSOPrediction}, an ENSO prediction model that achieves effective forecast skill at leads of up to 20 months, retaining its trained weights and gradient formulation. This long-lead skill rests on a convolutional feature extractor combined with a Transformer encoder--decoder, which predicts the Ni\~{n}o 3.4 index over the subsequent 24 months from a 12-month history of nine normalized ocean--atmosphere anomaly fields (Methods). Although sensitivities can be obtained for all nine fields, we focus on those to SST and surface wind stress, which most directly capture the coupled ocean--atmosphere processes underlying ENSO development. Whereas Chen et al. used these gradients to interpret an already skillful forecast~\cite{chen2025LongrangeENSOPrediction}, here we evaluate the sensitivities themselves, examining their statistical robustness, their quantitative correspondence with the model's responses to controlled input perturbations, and their consistency with established ENSO dynamics.

\begin{figure}[b]
\centering
\includegraphics[width=1.0\textwidth]{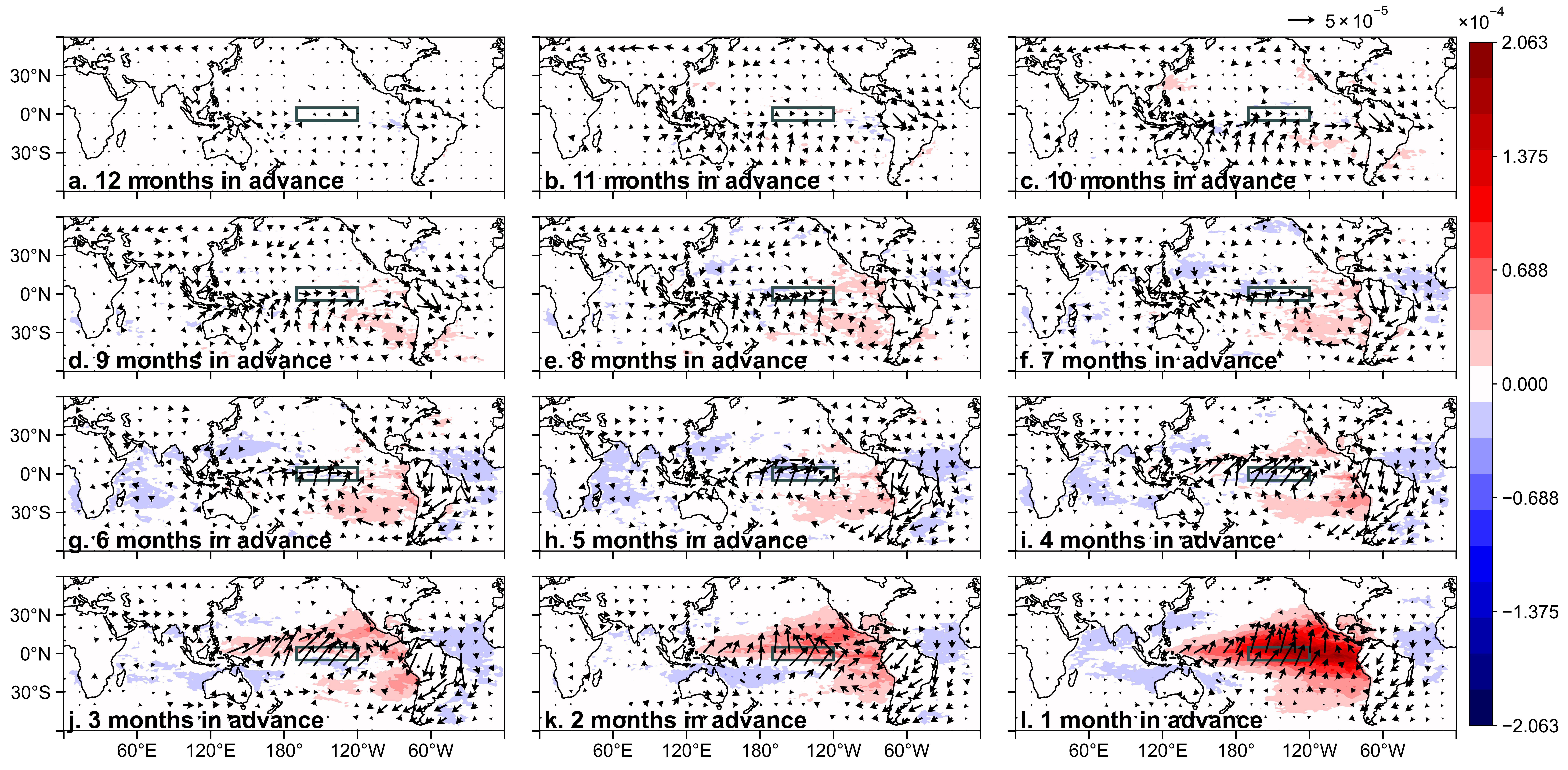}
\caption{\textbf{Sensitivity of the ENSO forecast metric to SST and wind-stress fields during El Ni\~{n}o events}. (a--l) Evolution of composite CTEFNet gradient-based sensitivities over the 12 months preceding the target periods of El Ni\~{n}o events. The SST and wind-stress components of the full multivariate sensitivity field are shown, with only the grid points significant at the 5\% level retained, based on a two-sided Student's $t$ test for the SST sensitivity and Hotelling's $T^{2}$ test for the zonal and meridional wind-stress sensitivities treated jointly. Shading indicates sensitivity to SST, and vectors represent sensitivities to zonal and meridional wind stress. The green box outlines the target Ni\~{n}o 3.4 region.}
\label{fig:Elnino-sensitivity}
\end{figure}

We compute sensitivities by backpropagating each predicted Ni\~{n}o 3.4 index to the preceding 12-month input sequence. For each El Ni\~{n}o event, gradients are averaged over a target period extending from November of the event year, when events typically approach their peak, through the subsequent months in which the observed index remains above 0.5. The resulting event gradients are then composited across 11 events during 1980--2021 (Methods). At each grid point, statistical significance is assessed using a two-sided Student's $t$ test for the SST sensitivity and Hotelling's $T^{2}$ test for the jointly considered zonal and meridional wind-stress sensitivities. All composites are masked to retain only the grid points significant at the 5\% level. Applying the same compositing and significance-testing procedures to La Ni\~{n}a events yields broadly sign-reversed patterns, with asymmetries in their seasonal evolution over the equatorial Pacific and in their inter-basin pathways (Fig.~S5). Leave-one-event-out recomposition further shows that the retained El Ni\~{n}o structures are not dominated by individual strong events and remain robust when the extreme 1982--83, 1997--98 and 2015--16 events are excluded in turn (Figs.~S6--S8). Finally, finite-perturbation experiments test whether these gradients predict the model's response. For each event and input month, SST and wind stress are jointly perturbed along their observed anomaly patterns at the significant grid points in the eastern tropical Pacific, North Tropical Atlantic and tropical Indian Ocean. Across all event--lead combinations in each region, the simulated Ni\~{n}o 3.4 response closely matches the first-order estimate obtained from the corresponding gradient (Text S4, Fig.~S9), confirming that the sensitivities remain quantitatively valid under finite-amplitude perturbations.

Building on the preceding evidence that these sensitivities are statistically robust and quantitatively consistent with the model's own responses, we next examine whether their spatiotemporal evolution is consistent with established ENSO dynamics. The full 12-month El Ni\~{n}o composite reveals a coherent precursor sequence (Fig.~\ref{fig:Elnino-sensitivity}). Sensitivities are weak 12--10 months in advance, although westerly wind-stress sensitivity is already confined to the tropical Pacific (Fig.~\ref{fig:Elnino-sensitivity}a--c). Coherent structures emerge more broadly from about 9 months before the target period (Fig.~\ref{fig:Elnino-sensitivity}d--f). Positive SST sensitivity in the tropical eastern Pacific, together with enhanced westerly wind-stress sensitivity over the tropical Pacific, aligns with the onset of the Bjerknes positive feedback~\cite{bjerknes1969ATMOSPHERICTELECONNECTIONSEQUATORIAL}. In this feedback, eastern-Pacific warming weakens the zonal temperature gradient and trade winds, promotes eastward warm-water transport, and further amplifies the initial warming. Concurrent negative SST sensitivity over the North Tropical Atlantic accords with a remote precursor pathway in which Atlantic cooling suppresses tropical Atlantic convection~\cite{ham2013SeaSurfaceTemperature}, shifts the Pacific Intertropical Convergence Zone (ITCZ), and weakens the Pacific trade winds, thereby favoring warm-water accumulation in the central and eastern Pacific~\cite{wu2017AtmosphericDynamicThermodynamica,alexander2002InfluenceENSOAirsea,park2022RoleClimatologicalNortha,jiang2021ImpactsTropicalNortha}. Enhanced southward wind-stress sensitivity over the tropical Atlantic further points to cross-equatorial heat transport and Walker-circulation adjustment as factors that precondition El Ni\~{n}o growth~\cite{chen2025LongrangeENSOPrediction}.

From spring to late summer (6--3 months in advance; Fig.~\ref{fig:Elnino-sensitivity}g--j), positive SST sensitivity expands westward into the central and western Pacific, while westerly wind-stress sensitivity strengthens over the western Pacific. This evolution matches canonical El Ni\~{n}o development, whereby growing SST anomalies further weaken the trade winds and amplify warming across the tropical Pacific~\cite{bjerknes1969ATMOSPHERICTELECONNECTIONSEQUATORIAL,timmermann2018NinoSouthernOscillationa,mcphaden2006ENSOIntegratingConcept}. Meanwhile, negative SST sensitivity over the North Tropical Atlantic strengthens and extends southward, consistent with Atlantic cooling suppressing convection and reorganizing the Walker circulation to induce tropical Pacific westerlies that favor El Ni\~{n}o development~\cite{wang2021JointBoostSupera}. Associated easterly wind-stress signals near Central and South America further reinforce the Atlantic--Pacific teleconnection~\cite{wang2006OverlookedFeatureTropical}. Negative SST sensitivity also emerges in the eastern Indian Ocean, consistent with an Indian Ocean Dipole-related pathway that promotes El Ni\~{n}o development through Walker-circulation adjustment~\cite{fan2024CouplingKeyTropicala,hameed2018ModelSuperNinos}.

Immediately before the event peak (2--1 months in advance; Fig.~\ref{fig:Elnino-sensitivity}k,l), tropical Pacific SST sensitivity strengthens further, while enhanced westerly wind-stress sensitivity points to eastward warm-water transport and accumulation in the central and eastern Pacific~\cite{mcphaden1999GenesisEvolution199798}. Negative SST sensitivity persists over the tropical Atlantic and is also widespread across the northern and equatorial Indian Ocean, the South China Sea, and the seas off northern Australia~\cite{nicholls1984SouthernOscillationIndonesian}. The Indian Ocean signal mirrors cooling that suppresses convection and induces divergence and westerly winds over the western tropical Pacific through a Matsuno--Gill response~\cite{matsuno1966QuasiGeostrophicMotionsEquatorial,gill1980SimpleSolutionsHeatinduced}. The resulting eastward-propagating downwelling Kelvin waves further reinforce El Ni\~{n}o development~\cite{xie2009IndianOceanCapacitora}.

Collectively, these statistically robust sensitivities are consistent with the canonical evolution of El Ni\~{n}o, showing that CTEFNet draws not only on local SST anomalies but also on remote SST forcing and wind-driven precursors that precondition equatorial Pacific warming. Together with their quantitative agreement with the model's responses to finite perturbations, this consistency with established ENSO dynamics supports the practical use of gradient-based sensitivity analysis for event diagnosis, precursor monitoring, and targeted observational design.

\section{Discussion}

DL forecast models are evaluated primarily by their predictive accuracy~\cite{liu2024EvaluationFiveGlobal}. For scientific applications, however, physical consistency is an equally important dimension of model quality that skill metrics alone cannot establish~\cite{geirhos2020ShortcutLearningDeep,reichstein2019DeepLearningProcess}. Model differentiability provides a direct means of probing this property. Gradients of a forecast objective with respect to the inputs reveal the dependencies across variables, regions and lead times that the model has actually learned. Here we establish gradient-based sensitivity analysis as a systematic evaluation framework. In an idealized convection benchmark, we directly validate learned sensitivities against those obtained from a physics-based adjoint model. In a realistic ENSO-prediction application, where no adjoint reference is available, we assess their statistical robustness, their agreement with the model's responses to finite perturbations, and their consistency with established physical mechanisms. Beyond improving model interpretability, this framework provides a physically grounded criterion for model evaluation that complements conventional accuracy-based metrics.

The Rayleigh--B\'enard benchmark shows that predictive accuracy and sensitivity fidelity are distinct properties of a differentiable surrogate. Although all four surrogates produce accurate and stable rollouts, their sensitivities for the same forecast objective differ substantially. Most notably, Swin Transformer achieves full-field accuracy second only to IFactFormer-RB yet agrees with the adjoint no better than FNO, the least accurate forecaster. This mismatch is consistent with the training objective: minimizing state-prediction error constrains the Jacobian of the learned forecast map only indirectly, even though this Jacobian determines the inferred sensitivities. Comparison across Fourier-operator and attention-based architectures further suggests that learned sensitivities reflect inductive biases not captured by conventional forecast metrics~\cite{tian2026EvaluatingMachineLearning}. Truncated or sparsified Fourier representations may be less effective at preserving localized perturbation pathways~\cite{li2021FourierNeuralOperator,guibas2022AdaptiveFourierNeural,kalimuthu2025LOGLOFNOEfficientLearning,koshizuka2024UnderstandingExpressivityTrainability}, whereas hierarchical shifted-window attention appears to encode long-range interactions in a Jacobian that remains spatially fragmented~\cite{liu2022SwinTransformerV2a,yang2021FocalSelfattentionLocalGlobal}. By contrast, the axis-wise kernel attention and implicit residual iteration of IFactFormer-RB may better preserve global coupling and the directional structure of gridded physical fields~\cite{li2023ScalableTransformerPDE,yang2025ImplicitFactorizedTransformer}. These mechanistic interpretations remain hypotheses, as controlled ablations would be required to separate architectural effects from parameter count, optimization dynamics and activation smoothness. The perturbation-ensemble experiment highlights the practical consequence of these differences: accurate deterministic forecasts do not guarantee faithful propagation of initial-condition uncertainty. Because the same Jacobian underlies both the sensitivities and the linear perturbation response, inaccurate sensitivities also lead to inaccurate responses to uncertain initial states. This connection extends the relevance of sensitivity fidelity beyond physical interpretation to inference and data assimilation. Predictive accuracy and sensitivity fidelity should therefore be treated as separate criteria in model evaluation.

The ENSO application demonstrates the value of gradient-based sensitivity analysis in a realistic prediction setting for which no matched adjoint reference is available. A single backward pass through CTEFNet reveals how the forecast depends on the preceding 12 months of multivariate ocean--atmosphere conditions. Obtaining comparable information using conventional approaches would require developing the adjoint of a coupled model or conducting large ensembles of perturbed forward integrations. Directly interpreting such gradients~\cite{chen2025LongrangeENSOPrediction} implicitly assumes that a skillful model has also learned faithful sensitivities, an assumption that the RB benchmark shows cannot be justified by predictive accuracy alone. We therefore assess the sensitivities themselves before drawing physical conclusions. They are statistically robust across events, stable in leave-one-event-out composites and quantitatively consistent with the model's response to finite-amplitude perturbations. Their spatiotemporal evolution is also consistent with canonical ENSO dynamics, including the development of the Bjerknes feedback and inter-basin influences from the tropical Atlantic and Indian Oceans. In the absence of an independent physical reference, this agreement constitutes consistency with established dynamics rather than direct validation. Nevertheless, it strengthens the evidence that CTEFNet's long-lead skill draws on dependencies of recognized physical relevance. More broadly, sensitivity fields identify the variables, regions and lead times on which individual forecasts depend, allowing trained networks to support precursor monitoring, event diagnosis and targeted observing strategies.

Taken together, the two case studies demonstrate that gradient-based sensitivity analysis provides a practical means of assessing whether DL forecast models encode physically consistent dependencies. Such assessment is increasingly important as DL models are used not only for prediction but also for scientific analysis and inference. When sensitivities underpin these applications, their fidelity warrants the same scrutiny as predictive skill. The gradients examined here are, however, local, first-order quantities that characterize model responses around a particular state or event composite rather than the full causal structure of the Earth system. Moreover, the present workflow can identify unfaithful sensitivities but does not itself correct them. An important direction for future work is therefore to determine whether derivative-informed and sensitivity-constrained training can improve sensitivity fidelity without compromising predictive skill when matched adjoint information is available. In more complex systems, where reliable reference sensitivities are difficult to obtain, the broader challenge is to design DL models with more stable and physically consistent gradients. This will require exploring architectures, training objectives, and regularization strategies that constrain local responses, potentially using information derived from finite-perturbation behavior and independent physics-based simulations without requiring explicit sensitivity targets.

More broadly, predictive validation of scientific neural networks establishes agreement in model outputs but does not by itself establish agreement in how those outputs respond to perturbations. When such models are intended to support scientific inference, optimization, or interpretation, their evaluation should therefore extend beyond state-space accuracy to the fidelity of the local response operators they encode.

\section{Methods}\label{sec4}

\subsection{Sensitivity analysis workflow and forecast models}

Following the framework outlined in Fig.~\ref{fig:sensitivity-framework}, we define the forecast maps used in the RB and ENSO applications. In the controlled Rayleigh--B\'enard (RB) benchmark, the model predicts the evolution of full temperature--velocity fields, allowing neural sensitivities to be evaluated against adjoint references. In the ENSO application, the model maps multivariate ocean--atmosphere histories to the future Ni\~{n}o 3.4 index, allowing sensitivities to be interpreted as precursor diagnostics.

In the RB benchmark, the forecast map is approximated by four differentiable surrogates: FNO~\cite{li2021FourierNeuralOperator} and AFNO~\cite{guibas2022AdaptiveFourierNeural}, which use global spectral mixing in Fourier space; Swin Transformer~\cite{liu2022SwinTransformerV2a}, which constructs long-range interactions through hierarchical shifted-window attention; and IFactFormer-RB, which combines axis-factorized kernel attention with implicit residual iteration~\cite{li2023ScalableTransformerPDE,yang2025ImplicitFactorizedTransformer} and is adapted here to the coupled temperature--velocity state. All four surrogates learn the same one-step evolution map and generate long-horizon forecasts through autoregressive rollout. They share identical data splits, a common relative $L_2$ loss and the same optimizer and learning-rate schedule, minimizing confounding from differences in data or training protocol when comparing their learned sensitivities.

Forecast accuracy for the four RB surrogates is quantified using the relative root-mean-square errors of the predicted state fields and the Nusselt number (Text S1). At each forecast lead and for each initialization window, the field error is defined as the $L_2$ norm of the difference between the predicted and reference states, computed across all grid points and physical channels and normalized by the $L_2$ norm of the reference state. The Nusselt-number error is defined analogously as the absolute prediction error normalized by the reference value. For each test trajectory, the squared relative errors are averaged across all initialization windows before taking the square root. The reported curves show the mean of these trajectory-level errors across the 20 test trajectories, with 95\% percentile confidence intervals obtained from 1{,}000 trajectory-level bootstrap resamples.

In the ENSO application, the forecast map instead predicts a scalar index: we adopt CTEFNet~\cite{chen2025LongrangeENSOPrediction}, a convolutional Transformer network that maps 12 consecutive months of multivariate ocean--atmosphere anomaly fields to the Ni\~{n}o 3.4 index over the subsequent 24 months. It combines a convolutional feature extractor, which captures regional precursor patterns through hierarchical spatial downsampling, with a Transformer encoder--decoder that models the cross-variable and temporal dependencies of ENSO evolution, and is trained with a mean-squared-error (MSE) loss on the predicted index.

Detailed training settings for the deep learning models, including hyperparameters, loss functions and computational cost, are reported in Text S2.

\subsection{Sensitivity objectives and gradient computation}

For a trained forecast model $\hat{Y}=f_{\theta}(X)$, the sensitivity calculation is specified by a scalar objective $J(\hat{Y})$, where $X$ denotes the input state or sequence. The corresponding input sensitivity field is
\begin{equation}
S_X=\frac{\partial J(\hat{Y})}{\partial X}.
\end{equation}
This field provides a local, first-order estimate of how perturbations to each input variable, grid location and, where applicable, input time affect the objective. Because $J$ is scalar, $S_X$ is obtained as a vector--Jacobian product that contracts the Jacobian of the learned forecast map with the output-space gradient defined by $J$, rather than reconstructing the full Jacobian. It therefore characterizes the model's response locally around the evaluated state and should not be interpreted as a complete causal description of the physical system. The gradient is computed by backpropagation through the trained model, without requiring separate perturbed forward integrations.

In the RB benchmark, the objective $J_{\mathrm{RB}}$ is the mean temperature within the target region at forecast lead step 20. Sensitivities are computed with respect both to the initial temperature and velocity fields and to the model state at each intermediate lead step by backpropagating through the remainder of the rollout. This procedure mirrors the backward-in-time propagation of the adjoint, ensuring that the surrogate and adjoint fields shown at each time represent gradients with respect to the state at that same time. The surrogate sensitivities are then evaluated against reference gradients from the adjoint RB model to determine whether accurate state prediction is accompanied by recovery of the physical dependency structure governing the target response.

Agreement with the adjoint is quantified for each trajectory using relative $L_2$ error and uncentered cosine similarity over the full sensitivity history preceding the objective time. Temperature and stacked velocity sensitivities are evaluated separately and then combined with equal weight. For each field $q\in\{\Theta,\boldsymbol{u}\}$, with $\hat{\boldsymbol{g}}_q$ and $\boldsymbol{g}^{\mathrm{adj}}_q$ denoting the surrogate and adjoint sensitivities, respectively, the relative error is
\begin{equation}
E_q=\frac{\lVert\hat{\boldsymbol{g}}_q-\boldsymbol{g}^{\mathrm{adj}}_q\rVert_2}{\lVert\boldsymbol{g}^{\mathrm{adj}}_q\rVert_2},
\end{equation}
where the norm spans all grid points and stored times. A value of zero indicates exact agreement. The cosine similarity at each stored time is
\begin{equation}
C_q(t)=\frac{\langle\hat{\boldsymbol{g}}_q(t),\boldsymbol{g}^{\mathrm{adj}}_q(t)\rangle}{\lVert\hat{\boldsymbol{g}}_q(t)\rVert_2\lVert\boldsymbol{g}^{\mathrm{adj}}_q(t)\rVert_2},
\end{equation}
and is averaged first over time and then equally across the temperature and velocity fields. Values of one, zero and minus one indicate perfect alignment, orthogonality and opposite directions, respectively. Boxplots summarize the 20 trajectory-level values; crosses denote their means, with 95\% percentile confidence intervals estimated from 1{,}000 trajectory-level bootstrap resamples.

For ENSO prediction, following Chen et al.~\cite{chen2025LongrangeENSOPrediction}, the objective is defined as the CTEFNet-predicted Ni\~{n}o 3.4 index for each target month associated with an El Ni\~{n}o event during 1980--2021. For each event, the target period begins in November of the El Ni\~{n}o year and extends through the subsequent months for which the observed index remains above 0.5. Gradients are computed for each target month with respect to the corresponding normalized 12-month input sequence, averaged first across target months within each event and then composited across events to produce sensitivity maps at each input lead month. Statistical significance is assessed using the 11 event-level sensitivity fields as independent samples. At each grid point, the mean SST sensitivity is tested against zero with a two-sided one-sample Student's $t$ test, whereas the zonal and meridional wind-stress sensitivities are tested jointly against a zero-mean vector with Hotelling's $T^{2}$ test. The composites are masked to retain only the grid points significant at the 5\% level. 

\subsection{Rayleigh--B\'enard benchmark dynamics}

The RB benchmark provides a controlled setting for comparing neural sensitivities with adjoint sensitivities from the governing equations. We solve the two-dimensional nondimensional Boussinesq equations at $\mathrm{Ra}=5\times10^{4}$ and $\mathrm{Pr}=1$ in a $[0,3]\times[0,1]$ domain with periodic horizontal boundaries, no-slip plates and fixed temperatures of 1 and 0 at the bottom and top, respectively. The equations are integrated using the Dedalus framework~\cite{burns2020DedalusFlexibleFramework} on a $128\times64$ grid with a time step of $10^{-2}$ up to $t=35$; reference sensitivities are obtained with its automated discrete-adjoint framework~\cite{skene2026FastAutomatedAdjoints}. States are stored every $0.1$ time units, corresponding to one forecast lead step of the surrogate models. Each trajectory begins from rest with an independently sampled, small-amplitude random temperature perturbation. We use 50 trajectories for training and 20 for testing. The surrogates operate on channel-wise standardized inputs and outputs, and their sensitivities are transformed back to the physical state variables via the chain rule, allowing direct comparison with the adjoint fields without additional rescaling or amplitude alignment. Further details of the equations, nondimensionalization and initialization are provided in Text S1.

\subsection{ENSO prediction data and preprocessing}

CTEFNet is trained on simulations from 18 CMIP6 models (Table S1), validated using ORAS5 ocean and ERA5 atmospheric reanalyses for 1958--1978, and evaluated independently using GODAS and ERA5 data for 1980--2021. Each sample comprises a 12-month sequence---the current and preceding 11 months---of sea surface temperature (SST), ocean heat content, mixed-layer depth, sea-surface salinity, sea-level pressure, zonal and meridional surface currents, and zonal and meridional wind stress, with the subsequent 24-month Ni\~{n}o 3.4 index as the prediction target. For each CMIP6 model and reanalysis dataset, monthly anomalies are calculated separately by removing the long-term trend and monthly climatology. All fields are then interpolated to a common $1^\circ \times 2^\circ$ grid spanning $60^\circ\text{S}$--$60^\circ\text{N}$ and all longitudes, and normalized using factors computed separately for the CMIP6 simulations and reanalyses. The resulting multivariate anomaly sequences are used for both forecasting and sensitivity analysis.

\section*{Data availability}
All datasets used in this work are available online. The Rayleigh--B\'enard convection data were generated using Dedalus, which is available at \url{https://github.com/DedalusProject/dedalus} (see Text~S1 for details). For ENSO prediction, the CMIP6 simulations are available at \url{https://metagrid.esgf-west.org/search}. The ORAS5 and ERA5 reanalysis data are available at \url{https://cds.climate.copernicus.eu/datasets/reanalysis-oras5?tab=overview} and \url{https://cds.climate.copernicus.eu/datasets/reanalysis-era5-single-levels-monthly-means?tab=download}, respectively. The GODAS data are available at \url{https://psl.noaa.gov/data/gridded/data.godas.html}.

\section*{Code availability}
The source code of this study is publicly available via GitHub at \url{https://github.com/QiChen-7-c/Sensitivity-Fidelity-Test}.

\begingroup
\useOriginalUrlSetting
\Urlmuskip=0mu
\bibliography{sensitivity.bib}
\endgroup

\section*{Acknowledgements}
This work is supported by the Digital Arabian Peninsula (DAP) Initiative funded by the office of the Vice President of Research at KAUST (REP/1/7475-01-01), National Natural Science Foundation of China (42276029), KAUST Competitive Research Grants (URF/1/6419-01-01) and Shenzhen Science and Technology Program (JCYJ20240813094213018).

\section*{Author contributions}
I.H. and P.Z. supervised the project. P.Z. and Q.C. conceived the idea. Q.C. designed and carried out the experiments, analysed the results and wrote the paper. H.C. reviewed and revised the manuscript. W.C. provided technical support for model development. All the authors discussed the study results.

\section*{Competing interests}
The authors declare no competing interests.

\end{document}